\documentclass[runningheads]{llncs}
\usepackage{graphicx,cite}
\usepackage{amsmath,amssymb} 
\usepackage{color}
\usepackage[width=122mm,left=12mm,paperwidth=146mm,height=193mm,top=12mm,paperheight=217mm]{geometry}

\newcommand{\etal}{{\emph{et~al.}}~}

\begin{document}
\pagestyle{headings}
\mainmatter

\title{Spatio-Temporal LSTM with Trust Gates \\for 3D Human Action Recognition} 

\titlerunning{Spatio-Temporal LSTM with Trust Gates for 3D Human Action Recognition}

\authorrunning{Jun Liu, Amir Shahroudy, Dong Xu, and Gang Wang}

\author{Jun Liu$^{\dagger}$, Amir Shahroudy$^{\dagger}$, Dong Xu$^{\ddagger}$, and Gang Wang$^{\dagger,}$\thanks{Corresponding author.}}


\institute{$\dagger$ School of Electrical and Electronic Engineering, Nanyang Technological University\\
$\ddagger$ School of Electrical and Information Engineering, University of Sydney\\
 \email{\{jliu029,amir3,wanggang\}@ntu.edu.sg} \ \ \ \   \email{dong.xu@sydney.edu.au}}

\maketitle

\begin{abstract}
3D action recognition \--- analysis of human actions based on 3D skeleton data \--- becomes popular recently due to its succinctness, robustness, and view-invariant representation.
Recent attempts on this problem suggested to develop RNN-based learning methods to model the contextual dependency in the temporal domain.
In this paper, we extend this idea to spatio-temporal domains to analyze the hidden sources of action-related information within the input data over both domains concurrently.
Inspired by the graphical structure of the human skeleton, we further propose a more powerful tree-structure based traversal method.
To handle the noise and occlusion in 3D skeleton data, we introduce new gating mechanism within LSTM to learn the reliability of the sequential input data and accordingly adjust its effect on updating the long-term context information stored in the memory cell.
Our method achieves state-of-the-art performance on 4 challenging benchmark datasets for 3D human action analysis.
\end{abstract}

\section{Introduction}
\label{sec:intro}


In recent years, action recognition based on the locations of major joints of the body in 3D space has attracted a lot of attention.
Different feature extraction and classifier learning approaches are studied for 3D action recognition \cite{presti20163d,han2016review,zhu2016handcrafted}.
For example, Yang and Tian \cite{eigenjointsJournal} represented the static postures and the dynamics of the motion patterns via eigenjoints and utilized a Na\"{i}ve-Bayes-Nearest-Neighbor classifier learning.
A HMM was applied by \cite{HOJ3D} for modeling the temporal dynamics of the actions over a histogram-based representation of 3D joint locations.
Evangelidis \etal \cite{skeletalQuads} learned a GMM over the Fisher kernel representation of a succinct skeletal feature, called skeletal quads.
Vemulapalli \etal \cite{vemulapalli2014liegroup} represented the skeleton configurations and actions as points and curves in a Lie group respectively, and utilized a SVM classifier to classify the actions.
A skeleton-based dictionary learning utilizing group sparsity and geometry constraint was also proposed by \cite{Luo_2013_ICCV}.
An angular skeletal representation over the tree-structured set of joints was introduced in \cite{hog2-ohnbar}, which calculated the similarity of these features over temporal dimension to build the global representation of the action samples and fed them to SVM for final classification.

Recurrent neural networks (RNNs) which are a variant of neural nets for handling sequential data with variable length, have been successfully applied to language modeling \cite{mikolov2011extensions,sundermeyer2012lstm,mesnil2013investigation}, image captioning \cite{vinyals2015show,xu2015show}, video analysis \cite{yue2015beyond,srivastava2015unsupervised,Singh_2016_CVPR,Jain_2016_CVPR,Alahi_2016_CVPR,Deng_2016_CVPR,Ibrahim_2016_CVPR,Ma_2016_CVPR,Ni_2016_CVPR,li2016online}, 
human re-identification \cite{varior2016siamese,varior2016learning},
and RGB-based action recognition \cite{donahue2015long,li2016action,wu2015ACMMM}.
They also have achieved promising performance in 3D action recognition \cite{du2015hierarchical,veeriah2015differential,nturgbd}.

Existing RNN-based 3D action recognition methods mainly model the long-term contextual information in the temporal domain to represent motion-based dynamics.
However, there is also strong dependency between joints in the spatial domain.
And the spatial configuration of joints in video frames can be highly discriminative for 3D action recognition task.


In this paper, we propose a spatio-temporal long short-term memory (ST-LSTM) network which extends the traditional LSTM-based learning to two concurrent domains (temporal and spatial domains).
Each joint receives contextual information from neighboring joints and also from previous frames to encode the spatio-temporal context.
Human body joints are not naturally arranged in a chain, therefore feeding a simple chain of joints to a sequence learner cannot perform well.
Instead, a tree-like graph can better represent the adjacency properties between the joints in the skeletal data.
Hence, we also propose a tree structure based skeleton traversal method to explore the kinematic relationship between the joints for better spatial dependency modeling.

In addition, since the acquisition of depth sensors is not always accurate, we further improve the design of the ST-LSTM by adding a new gating function, so called ``trust gate'', to analyze the reliability of the input data at each spatio-temporal step and give better insight to the network about when to update, forget, or remember the contents of the internal memory cell as the representation of long-term context information.

The contributions of this paper are: (1) spatio-temporal design of LSTM networks for 3D action recognition, (2) a skeleton-based tree traversal technique to feed the structure of the skeleton data into a sequential LSTM, (3) improving the design of the ST-LSTM by adding the trust gate, and (4) achieving state-of-the-art performance on all the evaluated datasets.

\section{Related Work}
\label{sec:relatedwork}

Human action recognition using 3D skeleton information is explored in different aspects during recent years \cite{actionletPAMI,7284883,MMMP_PAMI,MMTW,rahmani2014real,shahroudy2014multi,Wang_2016_CVPR,rahmani2015learning,Lillo_2016_CVPR,hu2016ECCV,chen_2016_icassp,rahmani20163d,liu2016IVC,cai2016TMM,al2016PRL,Tao_2015_ICCV_Workshops,shahroudy2016deep,du2016representation}.
In this section, we limit our review to more recent RNN-based and LSTM-based approaches.

HBRNN \cite{du2015hierarchical} applied bidirectional RNNs in a novel hierarchical fashion.
They divided the entire skeleton to five major groups of joints and each group was fed into a separated bidirectional RNN.
The output of these RNNs were concatenated to represent upper-body and lower-body, then each was fed into another set of RNNs.
The global body representation was obtained by concatenating the output of these two RNNs and it was fed to the next layer of RNN.
The hidden representation of the final RNN was fed to a softmax classifier layer for action classification.

Zhu \etal \cite{zhu2016co} added a mixed-norm regularization term to a deep LSTM network's cost function in order to push the network towards learning co-occurrence of discriminative joints for action classification.
They further introduced an internal dropout \cite{dropout} technique within the LSTM unit, which was applied on all the gate activations.

Differential LSTM \cite{veeriah2015differential} added a new gating inside LSTM to keep track of the derivatives of the memory states in order to discover patterns within salient motion patterns.
All the input features for each frame were concatenated and fed to the differential LSTM.

Part-aware LSTM \cite{nturgbd} separated the memory cell to part-based sub-cells and pushed the network towards learning the long-term context representations individually for each part.
The output of the network was learned over the concatenated part-based memory cells followed by the common output gate.

Unlike the above mentioned works, the framework proposed in this paper does not concatenate the joint-based input features, instead it explicitly models the dependencies between the joints and applies recurrent analysis over spatial and temporal domains concurrently. Besides, a novel trust gate is developed to make LSTM robust to noisy input data.

\section{Spatio-Temporal Recurrent Networks}
\label{sec:approach}


Human actions can be characterized by the motion of body parts over time.
In 3D human action recognition, we have three dimensional locations of the major body joints in each frame.
Recently, recurrent neural networks have been successfully employed for 
skeleton-based 3D action recognition \cite{du2015hierarchical,zhu2016co,nturgbd}.

Long Short-Term Memory (LSTM) networks \cite{lstm} are very successful extensions of the recurrent neural networks (RNNs).
They utilize the gating mechanism over an internal memory cell to learn and represent a better and more complex representation of the long-term dependencies among the input sequential data, thus they are suitable for feature learning over a sequence of temporal data.

In this section, first we will briefly review the standard LSTM networks, then describe the proposed spatio-temporal LSTM model and the skeleton-based tree traversal.
Next we will introduce an effective gating scheme for LSTM to deal with the measurement noise in the input data (body joint locations) for the task of 3D human action recognition.

\subsection{Temporal Modeling with LSTM}
\label{sec:approach:lstm}

A typical LSTM unit contains an input gate $i_t$, a forget gate $f_t$, an output gate $o_t$, and an output state $h_t$, together with an internal memory cell state $c_t$.
The LSTM transition equations are formulated as:

\begin{eqnarray}
\left(
   \begin{array}{ccc}
    i_{t} \\
    f_{t} \\
    o_{t} \\
    u_{t} \\
   \end{array}
\right)
&=&
\left(
   \begin{array}{ccc}
    \sigma \\
    \sigma \\
    \sigma \\
    \tanh \\
   \end{array}
\right)
\left(
   M
   \left(
       \begin{array}{ccc}
        x_{t} \\
        h_{t-1} \\
       \end{array}
   \right)
\right)\\
c_{t} &=& i_{t} \odot u_{t}  + f_{t} \odot  c_{t-1}
\label{eq:ct}\\
h_{t} &=& o_{t}  \odot \tanh( c_{t})
\label{eq:ht}
\end{eqnarray}
where $\odot$ indicates element-wise product, $x_t$ denotes the input to the network at time step $t$, and $u_t$ denotes the modulated input.
$\sigma$ is the sigmoid activation function.
$M: \Re^{D+d} \to \Re^{4d}$ is an affine transformation consisting of model parameters, where $D$ is the dimensionality of input $x_t$ and $d$ is the number of LSTM cell state units.

Intuitively, the input gate $i_{t}$ determines the extent to which the modulated input information $(u_{t})$ is supposed to update the memory cell at time $t$.
The forget gate $f_{t}$ determines the effectiveness of the previous state of the memory cell $(c_{t-1})$ on its current state $(c_{t})$. 
Finally, the output gate $o_{t}$ governs the amount of information output from the memory cell.
Readers are referred to \cite{graves2012supervised} for more details about the mechanism of LSTM.

\subsection{Spatio-Temporal LSTM}
\label{sec:approach:stlstm}
%
%


Very recent attempts on applying RNNs for 3D human action recognition \cite{du2015hierarchical,zhu2016co,veeriah2015differential,nturgbd} show outstanding performance and prove the strengths of RNNs in modeling the complex dynamics of the human actions in temporal space.



The main focus of these existing methods was on utilizing RNNs over temporal domain for discovering the discriminative dynamics and body motion patterns for 3D action recognition.
However, there is also discriminative information in static postures encoded within the joints' 3D locations in each individual frame and the sequential nature of skeleton data makes it possible to adopt RNN-based learning in spatial domain as well.
Unlike other existing methods, which concatenated the joints information, we extend the recurrent analysis towards spatial domain to discover the spatial dependency patterns between different joints at each frame.

\begin{figure}
	\begin{minipage}[b]{1.0\linewidth}
		\centering
		\centerline{\includegraphics[scale=.3]{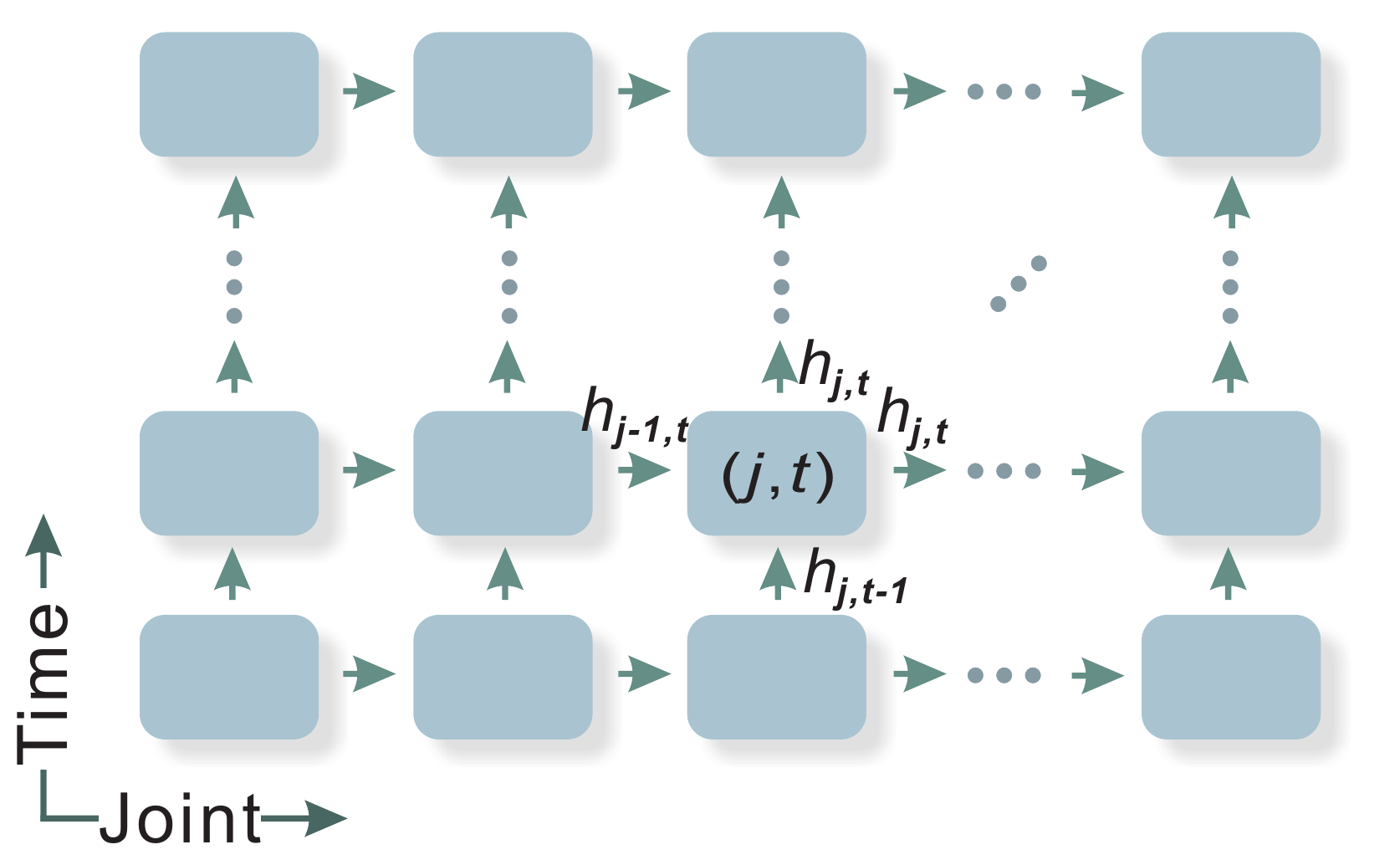}}
	\end{minipage}
	\caption{
		The illustration of the proposed spatio-temporal LSTM network.
		In the spatial direction, body joints in a frame are fed in a sequence. 
		In the temporal direction, the locations of the corresponding joints are fed over time.
		Each unit receives the hidden representation of previous joints and previous frames of the same joint as contextual information.}
	\label{fig:STLSTM}
\end{figure}

\begin{figure}
	\begin{minipage}[b]{0.25\linewidth}
		\centering
		\centerline{\includegraphics[scale=.23]{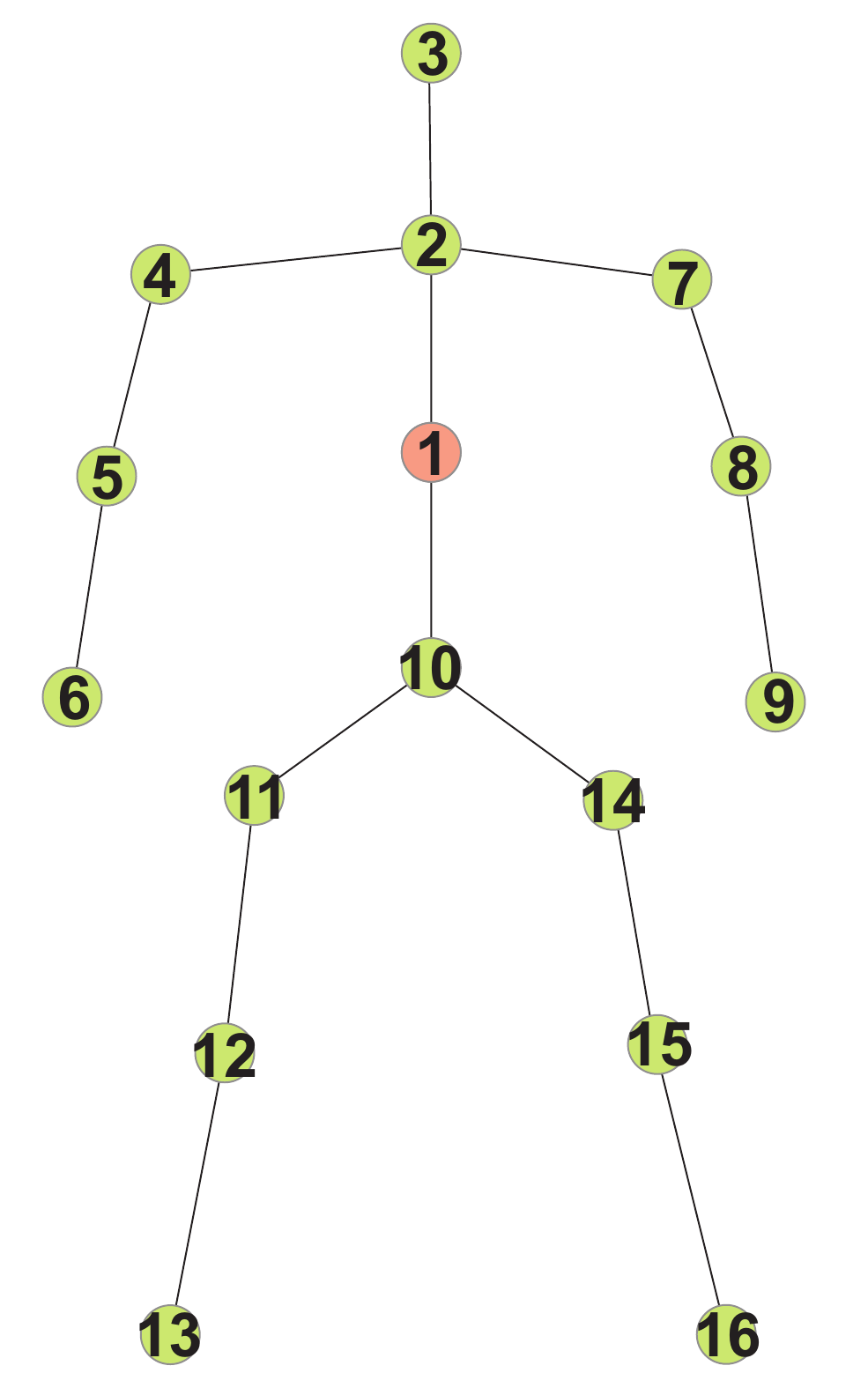}}
		\centerline{(a)}
	\end{minipage}
	\begin{minipage}[b]{0.36\linewidth}
		\centering
		\centerline{\includegraphics[scale=.22]{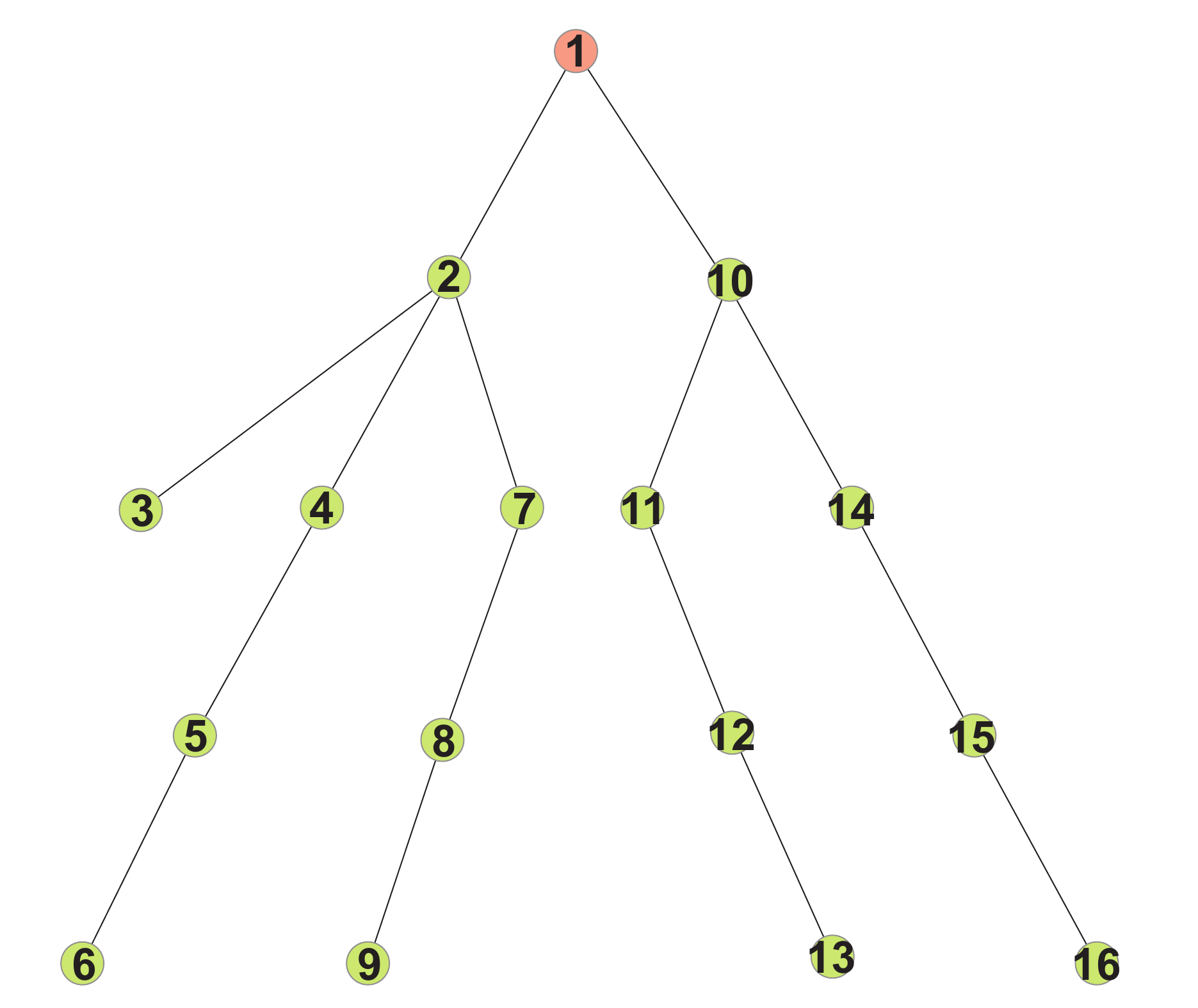}}
		\centerline{(b)}
	\end{minipage}
	\begin{minipage}[b]{0.36\linewidth}
		\centering
		\centerline{\includegraphics[scale=.22]{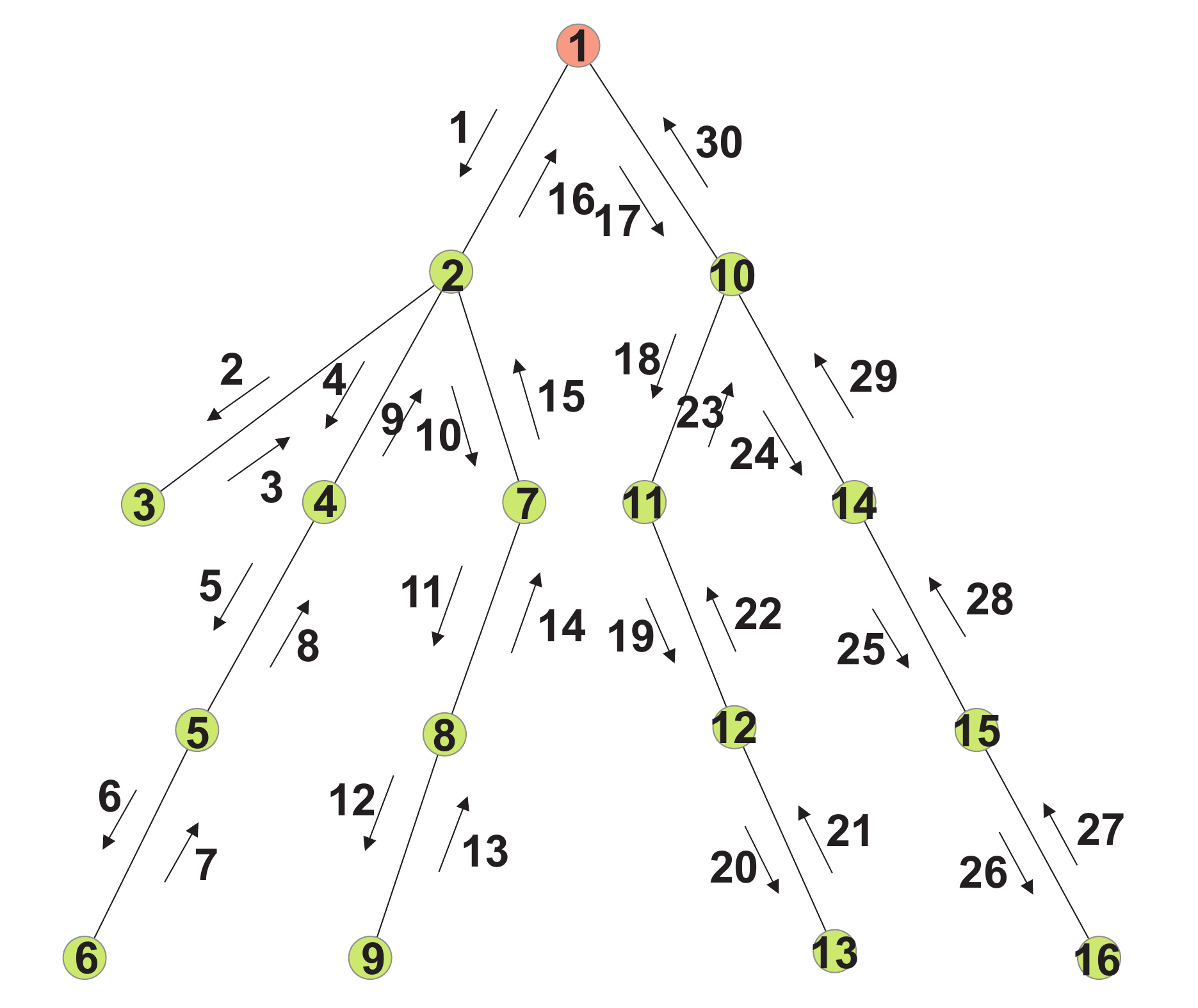}}
		\centerline{(c)}
	\end{minipage}
	\caption{(a) Skeletal joints of a human body. In the simple joint chain model, the joint visiting order is 1-2-3-...-16. (b) Skeleton is transformed to a tree structure. (c) Tree traversal over the spatial steps. The tree can be unfolded to a chain with the traversal, and the joint visiting order is 1-2-3-2-4-5-6-5-4-2-7-8-9-8-7-2-1-10-11-12-13-12-11-10-14-15-16-15-14-10-1.}
	\label{fig:tree16joints}
\end{figure}

In this fashion, we propose a spatio-temporal LSTM (ST-LSTM) model which simultaneously models the spatial dependencies of the joints and the temporal dependencies among the frames.
As shown in \figurename{ \ref{fig:STLSTM}}, every ST-LSTM unit corresponds to one of the skeletal joints.
Each of the units receives the hidden representation of the previous joint and also the hidden representation of its own joint from the previous frame.
In this section we assume joints are arranged in a chain-like sequence with the order shown in \figurename{ \ref{fig:tree16joints}(a)}.
In Section \ref{sec:approach:skeltree}, we will show a more advanced method to take advantage of the adjacency information of the body joints as a tree structure.

We use $j \in \{1,...,J\}$ and $t \in \{1,...,T\}$ to denote the indices of joints and frames respectively.
Each ST-LSTM unit is fed with its input ($x_{j, t}$, location of the corresponding joint at current frame), its own hidden representation at the previous time step $(h_{j,t-1})$, and the hidden representation of the previous joint at current frame $(h_{j-1,t})$.
Each unit is also equipped with two different forget gates corresponding to the two incoming channels of context information: $f_{j, t}^{S}$ for the spatial domain, and $f_{j, t}^{T}$ for the temporal domain.
The proposed ST-LSTM is formulated as:
\begin{eqnarray}
\left(
   \begin{array}{ccc}
    i_{j, t} \\
    f_{j, t}^{S} \\
    f_{j, t}^{T} \\
    o_{j, t} \\
    u_{j, t} \\
   \end{array}
\right)
&=&
\left(
   \begin{array}{ccc}
    \sigma \\
    \sigma \\
    \sigma \\
    \sigma \\
    \tanh \\
   \end{array}
\right)
\left(
   M
   \left(
       \begin{array}{ccc}
        x_{j, t} \\
        h_{j-1, t} \\
        h_{j, t-1} \\
       \end{array}
   \right)
\right)
\\
c_{j, t} &=&  i_{j, t} \odot u_{j, t} + f_{j, t}^{S} \odot  c_{j-1, t} + f_{j, t}^{T} \odot  c_{j, t-1}
\\
h_{j, t} &=& o_{j, t}  \odot \tanh( c_{j, t})
\end{eqnarray}



\subsection{Tree-Structure based Traversal}
\label{sec:approach:skeltree}

Arranging joints in a simple chain ignores the kinematic dependency relations between the joints and adds false connections between body joints which are not strongly related.
In human parsing, skeletal joints are popularly modeled as a tree-based pictorial structure \cite{zou2009automatic,yang2011articulated}, as illustrated in \figurename{ \ref{fig:tree16joints}(b)}.
In our ST-LSTM framework, it is also beneficial to model the spatial dependency of the joints based on their adjacency tree structure.
For example, hidden representation of the neck joint (number 2 in \figurename{ \ref{fig:tree16joints}(a)}) is expected to be more informative for the right hand joints (7,8,9) than the joint number 6.

However, trees cannot be directly fed into the ST-LSTM framework.
To mitigate this issue, we propose a bidirectional tree traversal method to visit joints in a sequence which maintains the adjacency information of the skeletal tree structure.

As illustrated in \figurename{ \ref{fig:tree16joints}(c)}, at the first spatial step, the root node (central spine joint) is fed to the network, then the network follows a depth-first traversal in the spatial domain. 
When it reaches a leaf node, it goes back.
In this fashion, each connection of the tree structure will be passed twice and the context information is fed along both directions.
Upon the end of the traversal, it gets back to the root node.

This traversal strategy guarantees the transmission of the data in both directions (top-down and bottom-up) inside the adjacency tree structure.
Therefore each node will have the contextual information from both its descendants and ancestors.
Compared to the simple chain model described in section \ref{sec:approach:stlstm}, this tree traversal technique can discover stronger long-term spatial dependency patterns based on the joints' adjacency structure.

In addition, the input to the ST-LSTM network at each step is limited to a single joint in a specific frame, which is much smaller in size compared to the concatenated input features of other existing methods.
As a result, we have much fewer model parameters and this can be considered as a weight sharing regularization inside our learning framework, which leads to better generalization in the scenarios with limited training samples.
This is an advantage in 3D action recognition, because most of the current datasets have a small number of training samples.

\begin{figure}
	\begin{minipage}[b]{0.99\linewidth}
		\centering
		\centerline{\includegraphics[scale=.328]{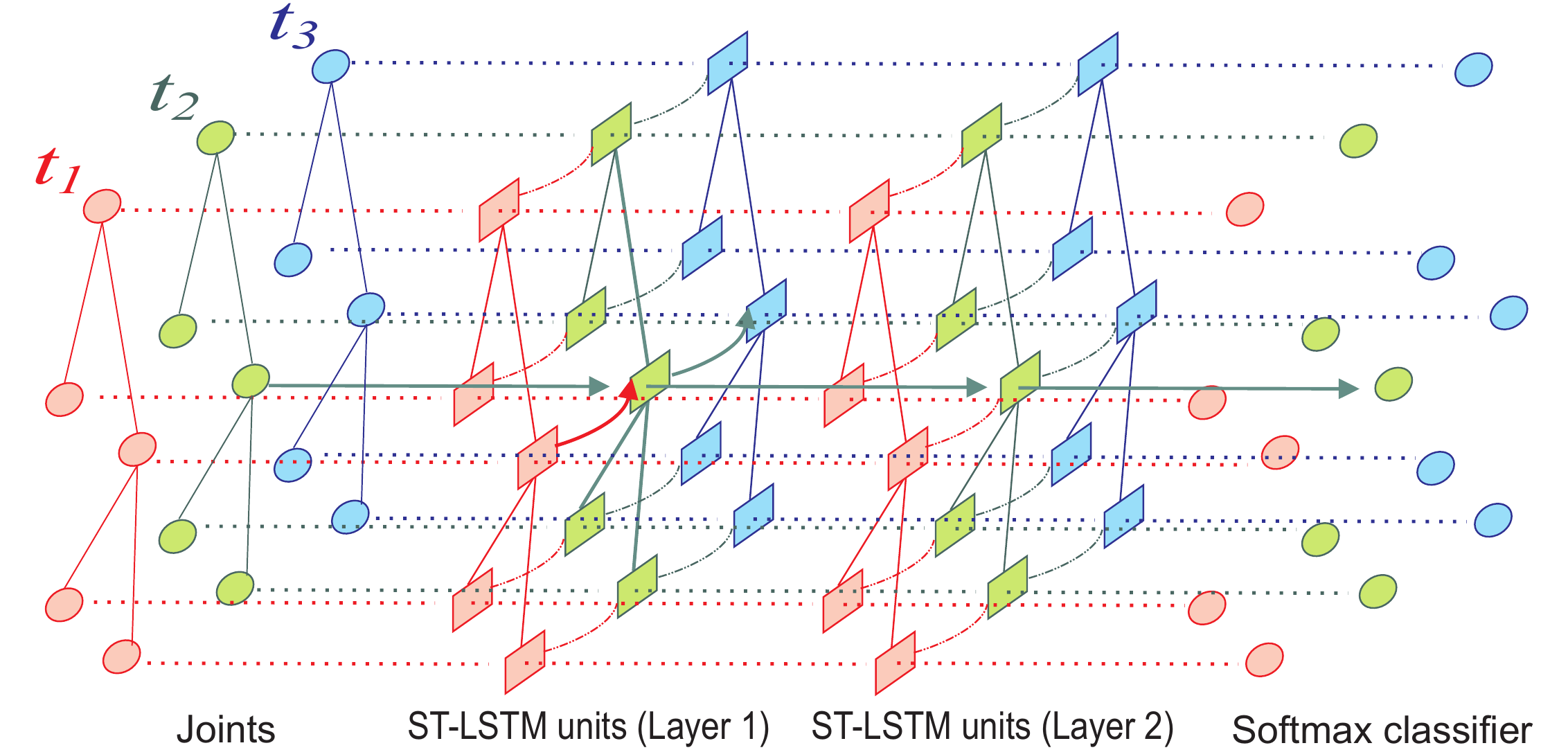}}
	\end{minipage}
	\caption{
A graphical model of the deep tree-structured ST-LSTM network.
For clarity, some arrows are omitted in the stacked network (better viewed in color).
In this figure, the output of the first ST-LSTM layer is fed to the second ST-LSTM layer as its input.
The second ST-LSTM layer's output is fed to softmax layer.
}
	\label{fig:stackedTreeSTLSTM}
\end{figure}

Similar to other LSTM implementations \cite{graves2013speechICASSP,sutskever2014sequence}, the representation capacity of our network can be improved by stacking multiple layers of the tree structured ST-LSTMs and constructing a deep yet completely tractable network, as illustrated in \figurename{ \ref{fig:stackedTreeSTLSTM}}.

\subsection{Spatio-Temporal LSTM with Trust Gates}
\label{sec:approach:trustgate}

%
%

The inputs of the proposed tree-structured ST-LSTM are the 3D positions of skeletal joints collected by sensors like Microsoft Kinect,
which are not always reliable due to noise and occlusion.
This limits the performance of the network.
To address this issue, we propose to add a new gate to the LSTM unit which analyzes the reliability of the input at each spatio-temporal step,
based on the estimation of the input from the available contextual information.

Our novel gating method is inspired by the works in natural language processing \cite{sutskever2014sequence} which predict next word based on LSTM representation of previous words.
This idea worked well because of the high dependency among the words in a sentence.
Similarly, since the skeletal joints often move together and this articulated motion follows common yet complex patterns at each spatio-temporal step, the input data $x_{j,t}$ is supposed to be predictable from the contextual representations $h_{j,t-1}$ and $h_{j-1,t}$.

This predictability inspired us to add new mechanism to ST-LSTM to predict the input and compare it with the actual incoming input.
The amount of the estimation error is used as input to a new ``trust gate''.
The derived trust value provides information to the long-term memory mechanism to learn better decisions about when and how to remember and forget the contents of the memory cell.
For example, when the trust gate finds out the current joint has wrong 3D measurements, it can block the input gate and prevent the memory cell from updating based on current unreliable input.

Mathematically, for an input at step $(j,t)$, we develop a function to generate its prediction, based on the available contextual information:
\begin{equation}
p_{j, t} = \tanh
\left(
   M_{p}
   \left(
       \begin{array}{ccc}
        h_{j-1, t} \\
        h_{j, t-1} \\
       \end{array}
   \right)
\right)
\end{equation}
where the affine transformation $M_p$ maps the data from $\Re^{2d}$ to $\Re^d$, so the dimensionality of $p_{j,t}$ is $d$.
\begin{figure}
	\centerline{\includegraphics[scale=0.42]{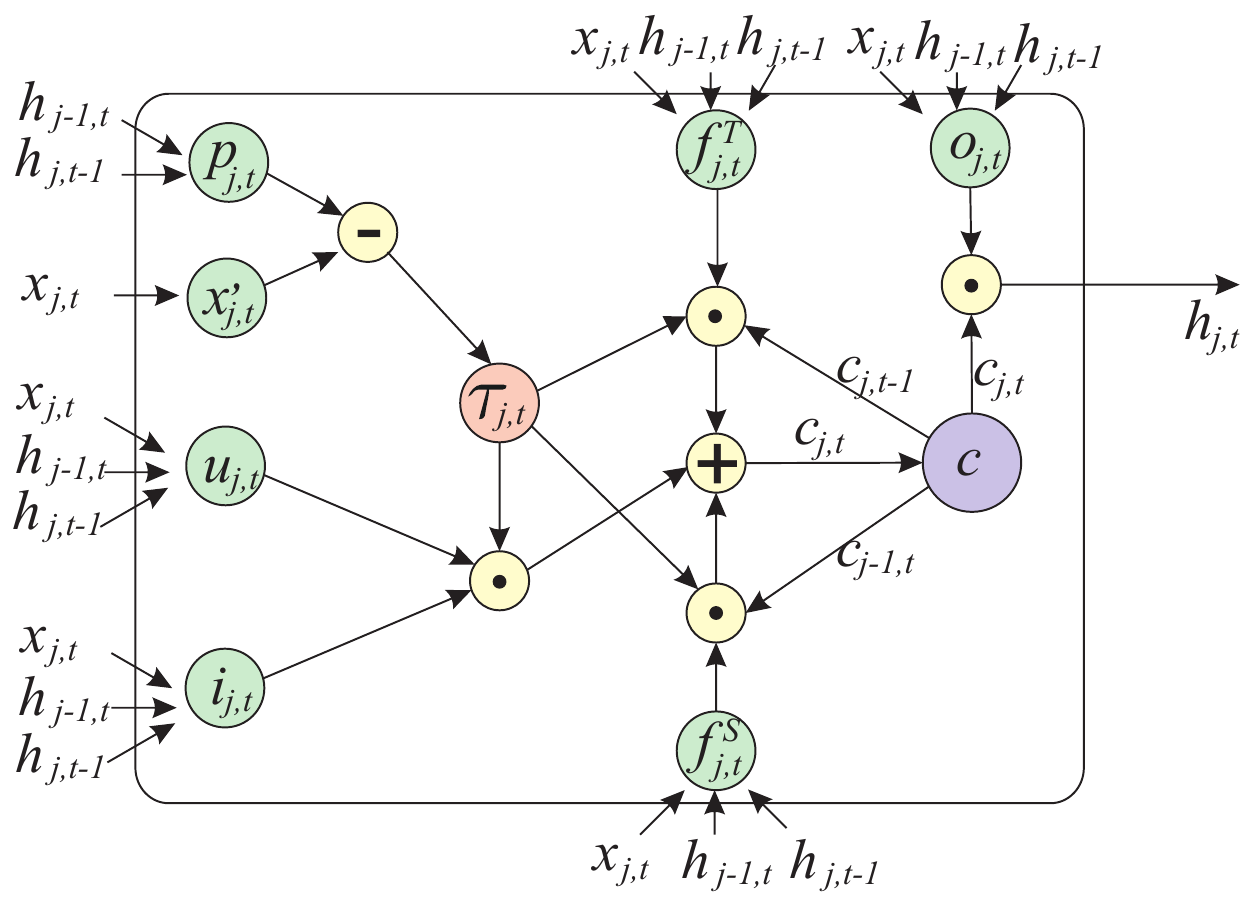}}
	\caption{Schema of the proposed ST-LSTM with trust gate.}
	\label{fig:TrustGateSTLSTMFig}
\end{figure}
It is worth noting that the contextual information at each step is not limited to the hidden states of the previous spatial step but it also includes the previous temporal step, i.e., the long-term memory information of the same joint in previous frames and the contextual information of other visited joints in the same frame are seamlessly incorporated.
Therefore, we can expect this function to be able to produce good predictions.

The activation of the proposed trust gate $\tau$ is a vector in $\Re^d$, which is similar to the activation of the input gate and the forget gate, and it will be calculated as:
\begin{eqnarray}
x'_{j, t} &=& \tanh
\left(
   M_{x}
   \left(
       \begin{array}{ccc}
        x_{j, t}\\
       \end{array}
   \right)
\right)
\\
\tau_{j, t} &=& G (x'_{j, t} - p_{j, t})
\label{eq:tau}
\end{eqnarray}
where $M_x: \Re^{D} \to \Re^{d}$ is an affine transformation, and the new activation function $G(\cdot)$ is an element-wise operation formulated as:
\begin{equation}
G(z) = \exp(-\lambda z^{2})
\end{equation}
In this equation, $\lambda > 0$ is a parameter to control the spread of the Gaussian function.
$G(z)$ produces a large response if $z$ is close to origin, and small response when $z$ has a large absolute value.

Utilizing the proposed trust gate, the cell state of the ST-LSTM neuron can be updated as:
\begin{equation}
c_{j, t} =  \tau_{j, t} \odot i_{j, t} \odot u_{j, t}
         + (\bold{1} - \tau_{j, t}) \odot f_{j, t}^{S} \odot  c_{j-1, t}
         + (\bold{1} - \tau_{j, t}) \odot f_{j, t}^{T} \odot  c_{j, t-1}
\end{equation}

If the new input $x_{j,t}$ cannot be trusted (because of noise or occlusion), then we need to take advantage of more history information and try to block the new input.
In contrast, if the input is reliable, we can let the learning algorithm update the memory cell by importing input information.

\figurename{ \ref{fig:TrustGateSTLSTMFig}} depicts the scheme of the new ST-LSTM unit empowered with the trust gate.
This can be learned similar to other gates by back-propagation.
The proposed trust gate technique is theoretically general and can be applied to other applications to deal with unreliable input data.

\subsection{Learning the Classifier}
\label{sec:approach:learning}
Since the action labels are always given at the video level, we feed them as the training outputs of the ST-LSTM at each spatio-temporal step.
The network learns to predict the action class $\hat{y}$ among a discrete set of classes $Y$ using a softmax layer.
The overall prediction of a video is computed by averaging the predictions of all the steps.
Empirically, this method provides better performance compared to the minimization of the loss at the last step only.

The objective function of our model is formulated as:
\begin{equation}
L = \sum_{j=1}^J \sum_{t=1}^T l(\hat{y}_{j,t}, y)
\end{equation}
where $l(\hat{y}_{j,t}, y)$ is the negative log-likelihood loss \cite{graves2012supervised} measuring the difference between the true label $y$ and the predicted result $\hat{y}_{j,t}$ at step $(j,t)$. The objective function can be minimized using back-propagation through time (BPTT) algorithm \cite{graves2012supervised}.

\section{Experiments}
\label{sec:exp}


The proposed model is evaluated on four datasets: NTU RGB+D dataset, SBU Interaction dataset, UT-Kinect dataset, and Berkeley MHAD dataset.
We conduct extensive experiments with different configurations as follows:

(1) ``ST-LSTM (Joint Chain)'': In this configuration, the joints are visited one by one in a simple chain order (see \figurename{ \ref{fig:tree16joints}(a)}).

(2) ``ST-LSTM (Tree Traversal)'': The proposed tree traversal strategy (\figurename{ \ref{fig:tree16joints}(c)) is adopted in this configuration to fully exploit the tree-based spatial structure of human joints.

(3) ``ST-LSTM (Tree Traversal) + Trust Gate'': This configuration involves the trust gate to deal with noisy input.

\subsection{Evaluation datasets}
\label{sec:exp:datasets}

{\bf NTU RGB+D Dataset} \cite{nturgbd}. To the best of our knowledge, this dataset is currently the largest depth-based action recognition dataset. It is collected by Kincet v2 and contains more than 56 thousand sequences and 4 million frames. A total of 60 different action classes including daily actions, pair actions, and medical conditions are performed by 40 subjects aged between 10 and 35. The 3D coordinates of 25 joints are provided in this dataset. The large intra-class and view point variations make this dataset very challenging. Due to the large amount of samples, this dataset is highly suitable for deep learning based action recognition.

{\bf SBU Interaction Dataset} \cite{yun2012two}. This dataset is captured with Kinect and contains 8 classes of two-person interactions. It includes 282 skeleton sequences in 6822 frames. Each skeleton has 15 joints. The challenges of this dataset include: (1) in most interactions, one person is acting and the other one is reacting; and (2) the joint coordinates in many sequences are of low accuracy.


{\bf UT-Kinect Dataset} \cite{HOJ3D}. This dataset contains 10 action classes performed by 10 subjects, captured with a stationary Kinect. Each action was performed twice by every subject. The locations of 20 joints are provided in this dataset. The high intra-class variation and viewpoint diversity makes it challenging.

{\bf Berkeley MHAD} \cite{ofli2013berkeley}. The MHAD dataset is captured by a motion capture system. It consists of 659 sequences and about 82 minutes of recording. Eleven different action classes were performed by 7 male and 5 female subjects. The 3D locations of 35 joints are provided in this dataset.

\subsection{Implementation details}
\label{sec:exp:impdetails}

%
%

In our experiments, each video sequence is divided to $T$ sub-sequences with the same length, and one frame was randomly selected from each sub-sequence.
Such a method adds randomness into the process of data generation and improves the generalization capability.
We observe this strategy achieves better performance in contrast to uniformly sampled frames.
We cross-validated the performance based on leave-one-subject-out protocol on NTU RGB+D dataset, and found $T=20$ as the optimum value.

We use Torch toolbox as the deep learning platform and an NVIDIA Tesla K40 GPU to run our experiments.
We train the network using stochastic gradient descent,
and set learning rate, momentum and decay rate as $2$$\times$$10^{-3}$, $0.9$ and $0.95$, respectively.
For our network, we set the neuron size $d$ to 128, and the parameter $\lambda$ used in $G(\cdot)$ to 0.5.
We use two ST-LSTM layers in the stacked network, and the applied probability of dropout is 0.5.
Though there are variations in terms of sequence length, joint number, and data acquisition equipment for different datasets, we use the same parameter settings mentioned above.
This indicates the insensitiveness of our method to the parameter settings, as it achieves promising results on all the datasets with the same configuration.

\subsection{Experimental results}
\label{sec:exp:res}



{\bf NTU RGB+D Dataset.} This dataset has two standard evaluation protocols \cite{nturgbd}. One is cross-subject evaluation, for which half of the subjects are used for training and the remaining are for testing.
The second is cross-view evaluation, for which two viewpoints are used for training and one is left out for testing.

\begin{table}[!hbp]
\caption{Experimental results (accuracies) on NTU RGB+D Dataset}
\label{table:resultNTU}
\centering
\scriptsize
\begin{tabular}{|l|c|c|}
\hline
Method & Cross subject & Cross view  \\
\hline
Lie Group \cite{vemulapalli2014liegroup}  & 50.1\% &  52.8\% \\
Skeletal Quads \cite{skeletalQuads} & 38.6\%  & 41.4\% \\
Dynamic Skeletons \cite{jianfang_CVPR15}  &  60.2\% & 65.2\%  \\
HBRNN \cite{du2015hierarchical}  & 59.1\% & 64.0\% \\
Part-aware LSTM \cite{nturgbd} & 	62.9\% &	70.3\%  \\
Deep RNN \cite{nturgbd} &  56.3\%  &  64.1\%  \\
Deep LSTM \cite{nturgbd} &   60.7\% & 67.3\%  \\
\hline
ST-LSTM (Joint Chain)  &	61.7\%	& 75.5\% \\
ST-LSTM (Tree Traversal) & 	65.2\% &	76.1\% \\
ST-LSTM (Tree Traversal) + Trust Gate & \textbf{69.2\%}	&  \textbf{77.7\%} \\
\hline
\end{tabular}
\end{table}

\begin{figure}
\begin{minipage}[b]{1.0\linewidth}
  \centering
  \centerline{\includegraphics[scale=0.23]{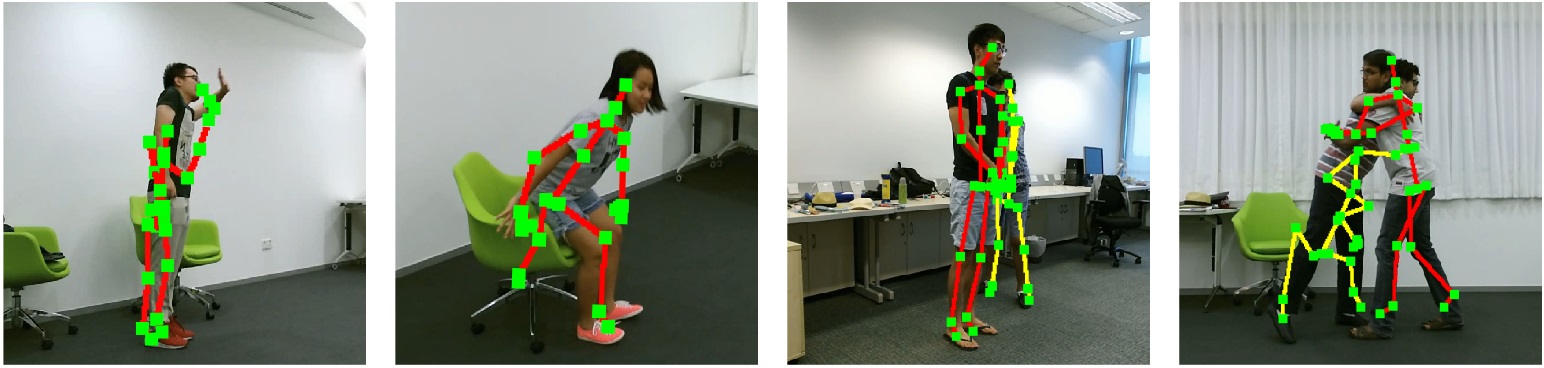}}
\end{minipage}
\caption{Example images with noisy skeletons from NTU RGB+D dataset.}
\label{fig:NTUNoisySamples}
\end{figure}

The results are shown in \tablename{ \ref{table:resultNTU}}.
Deep RNN and deep LSTM models concatenate the joints features at each frame and then feed them to the network to model the temporal dynamics and ignore the spatial dynamics.
As can be seen, both ``ST-LSTM (Joint Chain)'' and ``ST-LSTM (Tree Traversal)'' models outperform these methods by a notable margin.

It can also be observed that the trust gate brings significant performance improvement,
because the data acquired by Kinect is noisy and some joints are frequently occluded in this dataset.

A notable portion of samples of this dataset are captured from side view, and based on the design of Kinect's body tracking mechanism, side view skeletal data is less accurate than the front view.
To further show the effectiveness of trust gate, we analyze the performance using only the samples in side views.
When using ``ST-LSTM (Tree Traversal)'', the accuracy is 76.5\%, while ``ST-LSTM (Tree Traversal) + Trust Gate'' achieves 81.6\%.
This indicates the proposed trust gate can effectively handle severely noisy data.



To verify the effectiveness of layer stacking, we decrease the network size by using only one ST-LSTM layer, and the accuracies drop to 65.5\% (cross-subject) and 77.0\% (cross-view).
It indicates our two-layer stacked model has better representation strengths than a single-layer model.

The sensitivity of the proposed model to neural unit sizes and $\lambda$ values are also evaluated and the results are depicted in \figurename{ \ref{fig:NTUResultLambda}}.
When trust gate is used, our model achieves better performance for all the $\lambda$ values tested compared to the model without trust gate.

\begin{figure}
\begin{minipage}[b]{1.0\linewidth}
  \centering
  \centerline{\includegraphics[scale=.49]{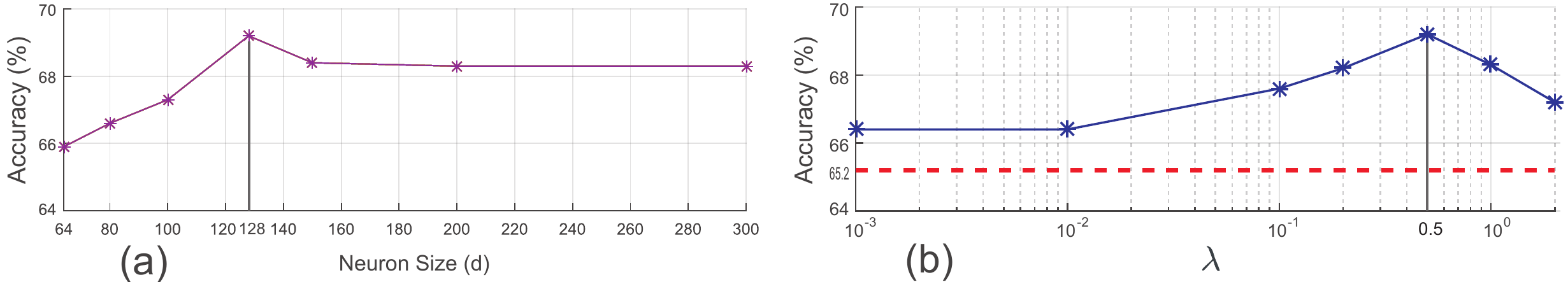}}
\end{minipage}
\caption{(a) Comparison of the performance for different neuron size ($d$) values on NTU RGB+D dataset (cross-subject). (b) Comparison of different $\lambda$ values on NTU RGB+D dataset (cross-subject). The blue line indicates the results when different $\lambda$ values are used for trust gate, and the red dashed line indicates the performance when trust gate is not added.}
\label{fig:NTUResultLambda}
\end{figure}

Finally, we evaluate the classification performance on early stopping conditions by feeding the first $p$ ($0<p<1$) portion of the testing video to the trained network on cross-subject protocol.
When setting $p$ as $0.1, 0.2, ..., 1.0$, the corresponding accuracies are $13.4\%$, $21.6\%$, $33.9\%$, $46.6\%$, $55.5\%$, $61.1\%$, $64.6\%$, $66.7\%$, $68.2\%$, $69.2\%$, respectively.
We can find that the results improve when a larger portion of video is fed.




{\bf SBU Interaction Dataset.}
We follow the standard experimental protocol of \cite{yun2012two} and perform 5-fold cross validation on SBU Interaction Dataset.
In this dataset, two human skeletons are provided in each frame, so our traversal visits the joints throughout the two skeletons over the spatial steps.
We summarize the results in terms of average classification accuracy in \tablename{ \ref{table:resultSBU}}.
In the table, \cite{zhu2016co} and \cite{du2015hierarchical} are both LSTM-based methods, which are more relevant to our model.

As can be seen, the proposed ``ST-LSTM (Tree Traversal) + Trust Gate'' model outperforms all other skeleton-based methods.
``ST-LSTM (Tree Traversal)'' yields higher accuracy than ``ST-LSTM (Joint Chain)'', as the latter adds some unreasonable links between the less related joints.

It is worth noting that deep LSTM \cite{zhu2016co}, Co-occurrence LSTM \cite{zhu2016co}, and HBRNN \cite{du2015hierarchical} all use the Svaitzky-Golay filter in temporal domain to smooth the skeleton joint positions to reduce the influence of the noise in the data captured by Kinect.
However, even without trust gate (which aims at handling noisy input), the ``ST-LSTM (Tree Traversal)'' model outperforms HBRNN and deep LSTM, and achieves comparable result (88.6\%) to Co-occurrence LSTM.
Once the trust gate is utilized, the accuracy jumps to 93.3\%.
We do not adopt any skeleton normalization operation, such as translation or rotation of the skeleton \cite{vemulapalli2014liegroup}, and achieve state-of-the-art performance.
We notice that \cite{lin2015deep} obtained very similar result (93.4\%) on SBU dataset. However, their method utilized both RGB and depth images, while our method just uses the skeleton data.

\begin{table}[t!]
	\parbox{0.45\linewidth}{
		\caption{Experimental results on SBU Interaction Dataset}
		\label{table:resultSBU}
		\centering
		\scriptsize
		\begin{tabular}{|l|c|}
			\hline
			Method & Accuracy   \\
			\hline
			Yun et al., \cite{yun2012two} & 80.3\% \\
			Ji et al., \cite{ji2014interactive} & 86.9\% \\
			CHARM \cite{li2015category} & 83.9\% \\
			HBRNN \cite{du2015hierarchical} (reported by \cite{zhu2016co}) & 80.4\% \\
			Co-occurrence LSTM \cite{zhu2016co} & 90.4\% \\
			Deep LSTM (reported by \cite{zhu2016co}) & 86.0\% \\
			\hline
			ST-LSTM (Joint Chain) & 84.7\% \\
			ST-LSTM (Tree) & 88.6\% \\
			ST-LSTM (Tree) + Trust Gate & \textbf{93.3\%} \\
			\hline
		\end{tabular}		
	}
	\hfill
	\parbox{0.45\linewidth}{
		\caption{Results on UT-Kinect Dataset (leave-one-out-cross-validation protocol \cite{HOJ3D})}
		\label{table:resultUTKinectprotocol1}
		\centering
		\scriptsize
		\begin{tabular}{|l|c|}
			\hline
			Method & Accuracy  \\
			\hline
			Histogram of 3D Joints \cite{HOJ3D} & 90.9\% \\
			Grassmann Manifold \cite{slama2015accurate} & 88.5\% \\
			Riemannian Manifold \cite{devanne20153} & 91.5\% \\
			\hline
			ST-LSTM (Joint Chain) & 91.0\% \\
			ST-LSTM (Tree) & 92.4\% \\
			ST-LSTM (Tree) + Trust Gate & \textbf{97.0\%} \\
			\hline
		\end{tabular}
	}
\end{table}


{\bf UT-Kinect Dataset.}
There are two popular protocols on UT-Kinect dataset. First is the leave-one-out-cross-validation protocol \cite{HOJ3D}.
Second is proposed in \cite{zhu2013fusing}, for which half of the subjects are used for training and the remaining are used for testing.
We use both protocols to evaluate the proposed method more extensively.

On the first protocol, our model achieves superior performance over other skeleton-based methods by a large margin, as shown in \tablename{ \ref{table:resultUTKinectprotocol1}}.
On the second evaluation protocol (\tablename{ \ref{table:resultUTKinectprotocol2}}),
our model achieves competitive result (95.0\%) to Elastic functional coding \cite{anirudh2015elastic} (94.9\%),
which is an extension of the Lie Group model \cite{vemulapalli2014liegroup}.


{\bf Berkeley MHAD.}
We follow the protocol in \cite{du2015hierarchical} on MHAD dataset, in which 384 sequences corresponding to the first 7 subjects are used for training and the 275 sequences of the remaining 5 subjects are used for testing.
The results are shown in \tablename{ \ref{table:resultMHAD}}.
Our method achieves the accuracy of 100\% without preliminary smoothing operations, which are adopted in \cite{du2015hierarchical}.




Besides, we have tested our model on {\bf MSR Action3D dataset} \cite{li2010action} following the protocol in \cite{du2015hierarchical}, and achieved an accuracy of 94.8\%, which is slightly superior to 94.5\% achieved by HBRNN \cite{du2015hierarchical}.

\subsection{Effectiveness of Trust Gate}
\label{sec:discussion}


To better study the effectiveness of the trust gate in the proposed network model, we specifically evaluate noisy samples from MSR Action3D dataset.
We manually rectify some noisy joints of these samples by referring to the corresponding depth maps, and compared the activations of the trust gates on noisy and rectified inputs.
As shown in \figurename{ \ref{fig:TrustGateEffect}(a)}, the activation of the trust gate is smaller when a noisy joint is fed, compared to the corresponding rectified joint.
This shows how the network reduces the impact of the noisy input data.

\begin{figure}[b!]
\begin{minipage}[b]{1.0\linewidth}
  \centering
  \centerline{\includegraphics[scale=.45]{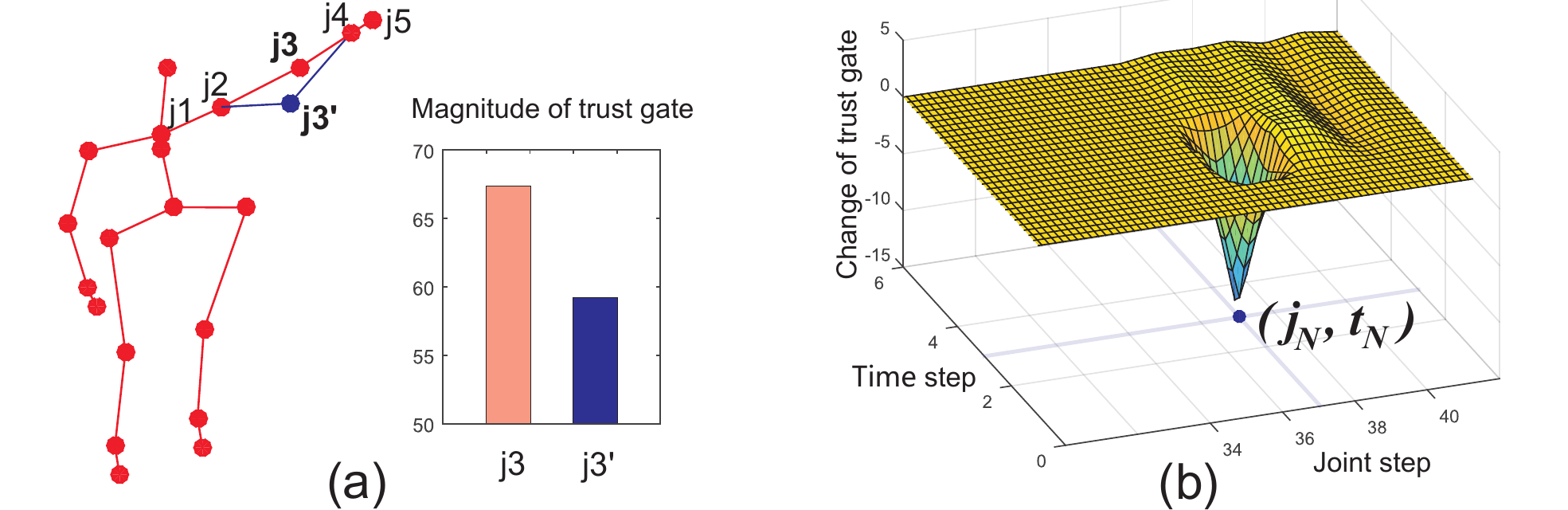}}
\end{minipage}
\caption{Behavior of trust gate when inputting noisy data. (a) $j_{3'}$ is a noisy joint location, and $j_3$ is the corresponding rectified joint position. In the histogram, the blue bar is the magnitude of the trust gate when inputting the noisy joint $j_{3'}$. The red bar is the magnitude of the corresponding trust gate when $j_{3'}$ is rectified to $j_3$. (b) The difference between the trust gate calculated when inputting the original data and that calculated when the noise is imposed at the $j_N$-th spatial step and $t_N$-th time step.}
\label{fig:TrustGateEffect}
\end{figure}

To comprehensively evaluate the trust gate, we also manually add noise to one joint for all testing samples on MHAD dataset. Note that MHAD dataset was captured with motion capture system, thus the skeleton joints are much more accurate than those collected by Kinect. 
We add noise to the right foot joint by moving the joint away from the original position. The direction of the translation vector is randomly chosen and the norm is also a random value around 30cm (this is a significant noise in the scale of human bodies). For each video, we add noise to the same joint at the same time step, and then analyze the effect in average.

We measure the difference in the magnitude of the trust gate activations between the original data and the noisy ones. For all the testing samples, we perform the same procedure, then calculate the average difference. The result is depicted in \figurename{ \ref{fig:TrustGateEffect}(b)}. We can see when the noisy data is fed to the network, the magnitude of the trust gate is reduced. This shows how the network ignores the noisy input, and tries to prevent it from affecting the network. In this experiment, we observe the overall accuracy does not drop after adding the noise.

\begin{table}[t!]
	\parbox{0.45\linewidth}{	
		\caption{Experimental results on UT-Kinect Dataset (half-vs-half protocol \cite{zhu2013fusing})}
		\label{table:resultUTKinectprotocol2}
		\centering
		\scriptsize
		\begin{tabular}{|l|c|}
			\hline
			Method & Accuracy \\
			\hline
			Skeleton Joint Features \cite{zhu2013fusing} & 87.9\% \\
			Lie Group \cite{vemulapalli2014liegroup} (reported by \cite{anirudh2015elastic}) & 93.6\%  \\
			Elastic functional coding \cite{anirudh2015elastic} & 94.9\% \\
			\hline
			ST-LSTM (Tree) + Trust Gate & \textbf{95.0\%} \\
			\hline
		\end{tabular}
	}
	\hfill
	\parbox{0.45\linewidth}{
		\caption{Experimental results on MHAD Dataset}
		\label{table:resultMHAD}
		\centering
		\scriptsize
		\begin{tabular}{|l|c|}
			\hline
			Method & Accuracy   \\
			\hline
			Vantigodi et al. \cite{vantigodi2013real} & 96.1\% \\
			Ofli et al. \cite{Ofli2014jvci} & 95.4\% \\
			Vantigodi et al. \cite{vantigodi2014action} & 97.6\% \\
			Kapsouras et al. \cite{kapsouras2014action} & 98.2\% \\
			HBRNN \cite{du2015hierarchical} & \textbf{100\%} \\
			\hline
			ST-LSTM (Tree) + Trust Gate & \textbf{100\%} \\
			\hline
		\end{tabular}
	}
\end{table}

\section{Conclusion}
\label{sec:conclusion}

In this paper we propose to extend the RNN-based 3D action recognition to spatio-temporal domain.
A new ST-LSTM network is introduced which analyses the 3D location of each individual joint in each video frame, at each processing step.
For better representation of the structured input to the network, a skeleton tree traversal algorithm is proposed which takes the adjacency graph of body joints into account and improves the performance of the network by arranging the most related joints together in the input sequence.
Due to the unreliability of the 3D input data, a new gating mechanism is also proposed to improve the robustness of the network against noise and occlusion.
The provided experimental results validate the proposed contributions and prove the effectiveness of our method by achieving superior performance over the existing state-of-the-art methods on four evaluated datasets.

\subsubsection*{Acknowledgement.}
The research is supported by Singapore Ministry of Education (MOE) Tier 2 ARC28/14, and Singapore A*STAR Science and Engineering Research Council PSF1321202099.
This research was carried out at the Rapid-Rich Object Search (ROSE) Lab at Nanyang Technological University.
The ROSE Lab is supported by the National Research Foundation, Singapore, under its Interactive Digital Media (IDM) Strategic Research Programme.
We also would like to thank NVIDIA for the GPU donation.

\bibliographystyle{splncs}
\bibliography{egbib}

\end{document}